\documentclass[conference]{IEEEtran}
\usepackage{times}
\usepackage{multirow}
\usepackage{makecell}
\usepackage{booktabs}
\usepackage{xcolor}
\usepackage{caption}
\usepackage{subcaption}

\usepackage[nopostdot,nogroupskip,nonumberlist]{glossaries}
\newglossaryentry{asp}{name=ASP, description={Assembly Sequence Planning},first={\textit{Assembly Sequence Planning} (ASP)}}
\newglossaryentry{auc}{name=AUC, description={Area Under Curve},first={\textit{Area Under Curve} (AUC)}}
\newglossaryentry{auroc}{name=AUROC, description={Area Under the Receiver Operating Characteristic Curve},first={\textit{Area Under the Receiver Operating Characteristic Curve} (AUROC)}}

\newglossaryentry{fpr}{name=FPR, description={False Positive Rate},first={\textit{False Positive Rate} (FPR)}}

\newglossaryentry{grace}{name=GRACE, description={Graph Assembly Processing Networks},first={\textit{Graph Assembly Processing Networks} (GRACE)}}

\newglossaryentry{id}{name=ID, description={In-Distribution},first={\textit{In-Distribution} (ID)}}

\newglossaryentry{mle}{name=MLE, description={Maximum Likelihood Estimation},first={\textit{Maximum Likelihood Estimation} (MLE)}}
\newglossaryentry{ml}{name=ML, description={Machine Learning},first={\textit{Machine Learning (ML)}}}

\newglossaryentry{nn}{name=NN, description={Neural Network},first={\textit{Neural Network} (NN)}}
\newglossaryentry{nf}{name=NF, description={Normalizing Flows},first={\textit{Normalizing Flows} (NF)}}

\newglossaryentry{oc-svm}{name=OC-SVM, description={One-class SVM},first={\textit{One-class SVM} (OC-SVM)}}
\newglossaryentry{ood}{name=OOD, description={Out-of-Distribution},first={\textit{Out-of-Distribution} (OOD)}}

\newglossaryentry{pdf}{name=PDF, description={Probability Density Function},first={\textit{Probability Density Function} (PDF)}}

\newglossaryentry{rasp}{name=RASP, description={Robotic Assembly Sequence Planning},first={\textit{Robotic Assembly Sequence Planning} (RASP)}}
\newglossaryentry{roc-auc}{name=ROC AUC, description={Area Under the Receiver Operating Characteristic Curve},first={\textit{Area Under the Receiver Operating Characteristic Curve} (ROC AUC)}}
\newglossaryentry{roc}{name=ROC, description={Receiver Operating Characteristic},first={\textit{Receiver Operating Characteristic} (ROC)}}

\newglossaryentry{tamp}{name=TAMP, description={Task and Motion Planning},first={\textit{Task and Motion Planning} (TAMP)}}
\newglossaryentry{tsne}{name=t-SNE, description={t-distributed Stochastic Neighbor Embedding},first={\textit{t-distributed Stochastic Neighbor Embedding} (t-SNE)}}
\newglossaryentry{tpr}{name=TPR, description={True Positive Rate},first={\textit{True Positive Rate} (TPR)}}

\usepackage[numbers]{natbib}
\usepackage{multicol}
\usepackage[bookmarks=true]{hyperref}
\usepackage{graphicx}
\newcommand{\linebreakand}{%
  \end{@IEEEauthorhalign}
  \hfill\mbox{}\par
  \mbox{}\hfill\begin{@IEEEauthorhalign}
}

\begin{document}

\title{Density-based Feasibility Learning with Normalizing Flows for Introspective Robotic Assembly}

\author{\authorblockN{Jianxiang~Feng\authorrefmark{1}\authorrefmark{2},
Matan~Atad\authorrefmark{1}\authorrefmark{3},
Ismael~Rodríguez\authorrefmark{2},
Maximilian~Durner\authorrefmark{2},\\
Stephan~Günnemann\authorrefmark{3} and
Rudolph~Triebel\authorrefmark{2}
}
\authorblockA{\authorrefmark{2}Institute of Robotics and Mechatronics, German Aerospace Center (DLR), 82110 Wessling, Germany}
\authorblockA{\authorrefmark{3}Department of Informatics, Technical University of Munich, 85748 Garching, Germany}
\authorblockA{\authorrefmark{1}Equal contribution. jianxiang.feng@dlr.de, matan.atad@tum.de}
}

\maketitle
\begin{abstract}
\gls{ml} models in \gls{rasp} need to be introspective on the predicted solutions, i.e. whether they are feasible or not, to circumvent potential efficiency degradation.
Previous works need both feasible and infeasible examples during training.
However, the infeasible ones are hard to collect sufficiently when re-training is required for swift adaptation to new product variants.
In this work, we propose a density-based feasibility learning method that requires only feasible examples. 
Concretely, we formulate the feasibility learning problem as Out-of-Distribution (OOD) detection with \gls{nf}, which are powerful generative models for estimating complex probability distributions. 
Empirically, the proposed method is demonstrated on robotic assembly use cases and outperforms other single-class baselines in detecting infeasible assemblies.
We further investigate the internal working mechanism of our method and show that a large memory saving can be obtained based on an advanced variant of \gls{nf}.
\end{abstract}

\IEEEpeerreviewmaketitle

\section{Introduction}
To embrace the trend of shorter product life cycles and greater customization, \gls{rasp} empowered with \gls{ml} models for productivity enhancement has received more attention over the past years ~\cite{atad2023efficient, rodriguez2020pattern, ma2022planning, Zhao2019}.
However, \textit{data-driven} models are reported to behave unreliably with inputs differing from the training distribution, e.g., assemblies with distinct customization~\cite{sinha2022system}. 
In other words, the assembly robot is \textit{unaware} of the predicted solution's feasibility, which requires an intrinsic understanding of the geometry of assemblies and the capability of the robotic system~\cite{rodriguez2019iteratively}.
This introspective capability is essential for learning-enabled robots to adapt their knowledge and avoid catastrophic consequences~\cite{feng2022introspective}. 
The lack of introspection in \gls{rasp} can lead to prolonged planning time induced by re-planning after failed execution of an infeasible plan.
To address this issue, feasibility learning has been studied~\cite{wells2019learning, driess2020deep, xu2022accelerating, yang2022sequence, atad2023efficient} based on a setting with \textit{infeasible assemblies included}.
We argue that this setting is undesirable in practice because of the risk of incomplete coverage of all possible infeasible cases and high time costs for generating sufficient infeasible training cases.
These aggravate the situation when flexible and efficient adaptation across different product variants is required.

To establish introspection for assembly robots with \textit{only feasible assemblies} in mind, we seek to model the feasibility of an assembly with \gls{nf}, which are a powerful class of generative models excelling at density estimation~\cite{dinh2016density}.
Concretely, we train the \gls{nf} model with \gls{mle} based on \textit{feasible assemblies alone} to estimate the density of \gls{id} data, i.e. feasible assemblies. Hence, infeasible assemblies can be detected via a lower predicted likelihood as \gls{ood}.
 
\begin{figure}[t]
	\centering
	\includegraphics[width=1.\linewidth]{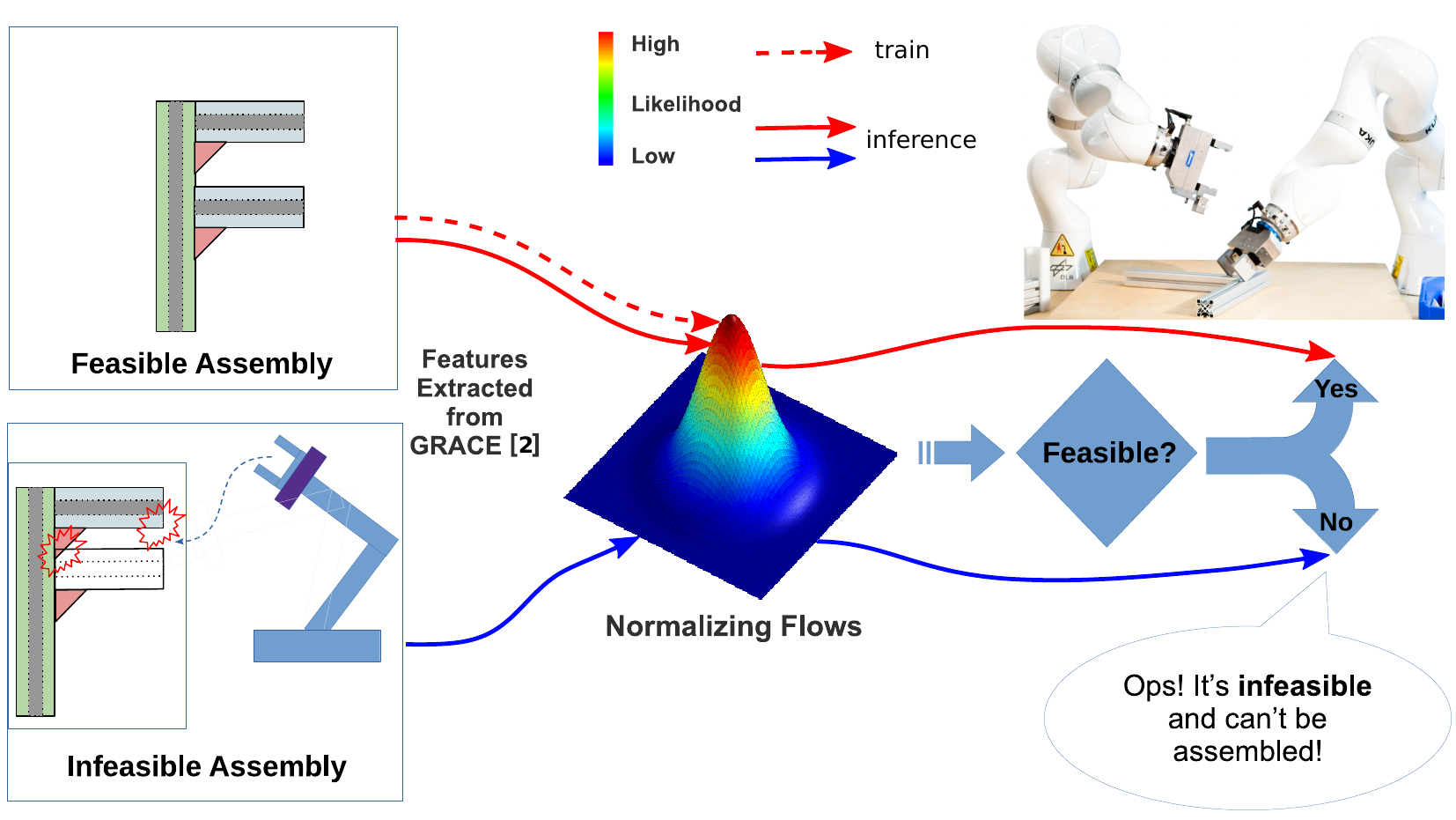}
	\caption{\textbf{Overview of the proposed method} on an assembly scenario with a dual-armed robotic system (used in our setting). The distribution of feasible assemblies is modeled during training with \gls{nf}. In test time, infeasible assemblies are identified by their low-likelihood.}
	\label{fig:teaser}
\end{figure}
We examine the proposed idea in a robotic assembly use case, in which different types of aluminum profiles are assembled with a dual-armed robot to create target structures (see Fig.~\ref{fig:teaser}). 
We collected assembly data in simulation and trained the \gls{nf} on features of \textit{only feasible} assemblies extracted from the \gls{grace} proposed in~\cite{atad2023efficient}.
The \gls{nf} model is then used to predict the likelihood of test data which includes both feasible and infeasible assemblies.
As we learn the feasibility by estimating the density of feasible cases, the predicted outputs from \gls{nf} represent how likely the given assemblies are feasible.
Based on a threshold selected on a validation set, we can then detect infeasible assemblies. Empirically, we demonstrate better results with the proposed method against other baselines on detecting infeasible assemblies in terms of \gls{auroc} in the setting where only feasible assemblies are available.
We further investigate the major contributing factors of \gls{nf} and significantly decrease the memory costs (i.e., number of network layers) by employing a more elaborate base distribution~\cite{stimper2022resampling}.
 
\section{Related Work}

\subsection{Feasibility Learning}
The major body of work on feasibility learning is concentrated on plan or action feasibility learning in TAMP, while our goal is to learn the feasibility of assemblies directly by distilling the knowledge of assembly geometry and capability of the robot system.
\citet{wells2019learning} trained a feature-based SVM model to directly predict the feasibility of an action sequence based on experience, which is hard to scale to scenarios with different numbers and types of objects. 
\citet{driess2020deep} and a recent follow-up \cite{xu2022accelerating} predict if a mixed-integer program can find a feasible motion for a required action based on visual input. 
Besides, \citet{yang2022sequence} predict a plan's feasibility with a transformer-based architecture using multi-model input embeddings.  \citet{atad2023efficient} introduced GRACE, a graph-based feature extractor for assemblies, capable of identifying infeasible assemblies when trained with both feasible and infeasible cases.
Different from us, these methods work in a two-class setting, requiring failing action sequences to be included in the training set and then use binary feasibility classifiers.

\subsection{Normalizing Flows for Out-of-Distribution Detection}
\gls{nf}~\cite{dinh2014nice} are a family of deep generative models with expressive modeling capability for complex data distributions where both sampling and density evaluation can be efficient and exact. 
Among a diverse set of flow architectures, Affine Coupling Flows~\cite{dinh2016density} have gained huge popularity for their scalability to big data with high dimensionality and efficiency for both forward and inverse evaluation. 
These merits make \gls{nf} more practically advantageous for \gls{ood} detection~\cite{kirichenko2020normalizing} when compared with other more principled but run-time inefficient uncertainty estimation methods~\cite{pmlr-v119-lee20b}.
In the context of task-relevant \gls{ood} detection, the practice of PostNet~\cite{charpentier2020posterior} of operating on feature embeddings, provides a more reasonable modeling ability.
The potentials of \gls{nf} for \gls{ood} detection have been demonstrated in other domains~\cite{rudolph2021same, zhang2020hybrid}, inspiring us to use them for feasibility learning.

\section{Method}
\subsection{Problem Setting}
Our goal is to predict the feasibility of assemblies relying only on feasible ones by formulating the problem as an \gls{ood} detection.
Given a data-set $\mathcal{D}$ of $N$ feature embeddings of feasible assemblies $\{\textbf{a}_i\}_{i=1}^{N}$, where $\textbf{a}_i \in \mathcal{R}^h$ is drawn from an unknown distribution $P_{feasible}$ with \gls{pdf} $p_{f}$, a density estimator, denoted by $q_{\theta}: \mathcal{R}^h \rightarrow \mathcal{R}$, approximates the true $p_{f}$ with \gls{mle} for its parameters $\theta$ based on $\mathcal{D}$. 
During inference, given a threshold $\delta \in \mathcal{R}$, the feature of a test assembly $\hat{\textbf{a}}_i$ is classified as \gls{ood}, i.e. infeasible, if $q_{\theta}(\hat{\textbf{a}}_i) < \delta$, otherwise as \gls{id}, i.e. feasible.

\subsection{Density-based Learning with NF}
In this work, \gls{nf} are used to estimate the density of feasible assemblies.
\gls{nf}, denoted by $f_{\theta}: \mathcal{R}^h \rightarrow \mathcal{R}^h$, are defined by a chain of \textit{diffeomorphisms} (invertible and differentiable mappings) that transform a base distribution $p(\textbf{z}),\ \textbf{z} \in \mathcal{R}^h$ (e.g. an isotropic Gaussian) to the data distribution $q_{\theta}$ (in our case $p_{f}$).
Based on the Change-of-Variables formula, the likelihood of an embedding of an assembly is obtained by
\begin{align} 
    q_{\theta}(\textbf{a}) &= p( f_{\theta}^{-1} (\textbf{a}) )\left\vert \det \left( \frac{\partial f_{\theta}^{-1} (\textbf{a})}{\partial \textbf{a}}\right) \right\vert \label{eq:change_of_variable} 
\end{align}
$\theta$ is optimized with \gls{mle} based on feasible data only, where the log likelihood is defined as:
\begin{align} 
    \log{q_{\theta}(\textbf{a})} = \log{p( f_{\theta}^{-1} (\textbf{a}) )} + \log{\left\vert \det \left( \frac{\partial f_{\theta}^{-1} (\textbf{a})}{\partial \textbf{a}}\right) \right\vert}
    \label{eq:mle} 
\end{align}
To this end, the inverse flow $f^{-1}$ and the log determinant of the Jacobian need to be tractable and efficient. 
We employ the Real-NVP \cite{dinh2016density} that is composed of multiple layers of affine coupling flows. 
As the input to the \gls{nf}, a data-set of feature embeddings for feasible assemblies $\mathcal{D}$ is extracted from a pre-trained \gls{grace}~\cite{atad2023efficient}, which represents each assembly structure as a graph of its parts and their respective surfaces. To create a single feature embedding per assembly, a channel-wise mean pooling is applied on the graph's part nodes. Different to previous works, the dimension of this embedding is independent of the number of assembly parts. 

During inference, given a test assembly embedding, the trained \gls{nf} $q_{\theta}$ predicts a log-likelihood score and determines its feasibility based on a pre-defined threshold $\delta$, which we selected with a validation set.

\section{Experiments}

\subsection{Data-set}
We applied our in-house simulation software MediView to randomly generate synthetic assemblies, each with $5$ or $6$ aluminum parts.
The software was tasked with putting together these structures with brute-force search while considering geometry restrictions and those imposed by the capabilities of the dual-armed robotic system \emph{KUKA LBR Med} (seen in Fig.~\ref{fig:teaser}). We label structures that were successfully assembled as feasible and ones for which the software failed as infeasible. 
The resulting data-set consists of $6036$ $5$-parts and $2865$ $6$-parts assemblies. For the training set, we used feasible-labeled assemblies alone. The validation and testing sets were balanced with both feasible and infeasible assemblies\footnote{This is still a single-class training setting since the validation set is only used for model selection.}. 

\subsection{Implementation Details}
We pre-trained \gls{grace} \cite{atad2023efficient} with its default parameters to retrieve a $94$-dimensions embedding per assembly. We implemented the \gls{nf} model using \cite{normflows} and experimented with Gaussian and Resampling \cite{stimper2022resampling} base distributions\footnote{Code and training data are available at \url{https://github.com/DLR-RM/GRACE}.}. For training the \gls{nf}, we chose a batch size of $32$ and a learning rate of $1e-5$ with Adam optimizer. The number of coupling flows was chosen with hyper-parameter search on a validation set. Each affine coupling flow contained $4$ layers with $94$ hidden channels per layer. 

We measure the separation between the feasibility classes with the binary classification metrics \gls{fpr} and \gls{tpr} to derive an \gls{auroc} score. In this setting, a positive instance is a feasible assembly and a negative an infeasible one.

\subsection{Results}

In Table~\ref{tab:classifiers}, we compare our method to baselines on predicting the feasibility of 5- and 6-part assemblies. 
The \gls{nf} model with Gaussian base distribution achieves the highest score with a deep $749$-layered network, outperforming the \gls{oc-svm}~\cite{scholkopf1999support} and the naive \gls{grace}~\cite{atad2023efficient}. In this setting, \gls{grace}, trained on \textit{feasible assemblies only}, predicts an assembly sequence for a test instance and infer the assembly's feasibility based on the success of its sequencing process.
More practically relevant, the \gls{nf} variant with the more expressive Resampling base distribution \cite{stimper2022resampling} can reach comparably good results with a much smaller network (109 vs. 749 layers).
This benefit of memory efficiency is highly relevant for robotic systems with only restricted computation resources (e.g., mobile manipulators). 
Contrary to \gls{grace}'s sequencing process, we only require a single-pass through the feature-extraction pipeline, independent of the size of the assembly, and could therefore determine the feasibility of  multiple batched assemblies at once.

\begin{table}[t]
  \centering
  \begin{tabular}{l c c}
    \toprule
    \multirow{2}{*}{Classifier} & \multicolumn{2}{c}{\gls{auroc} ($\uparrow$)} \\
                                & $5$-parts & $6$-parts \\
    \midrule
      \gls{grace} $+$ \gls{nf}, Gaussian dist., 749 layers (ours) & \textbf{0.85} & \textbf{0.83} \\
      \gls{grace} $+$ \gls{nf}, Resampling dist., 109 layers  (ours) & 0.83 & - \\
      \gls{oc-svm}~\cite{scholkopf1999support} & 0.74 & 0.59 \\
      \gls{grace} \cite{atad2023efficient}, feasible-only setting & 0.61 & 0.57 \\
    \bottomrule
  \end{tabular}
    \caption{Feasibility classifiers \gls{auroc} score on  balanced test sets of 5- and 6-part assemblies.}
    \label{tab:classifiers}
\end{table}

\subsection{Discussion}
For an insight into how \gls{nf} works on feasibility learning, we study the impacts of the flow transformations from the perspectives of two quantities: 1.~likelihoods; 2.~sample coordinates.
While the former represents the density estimation ability of \gls{nf}, the latter provides us a hint on how \gls{nf} shifts the samples from the flow input space into its latent space. 

\begin{figure}
    \centering
    \includegraphics[width=1\linewidth]{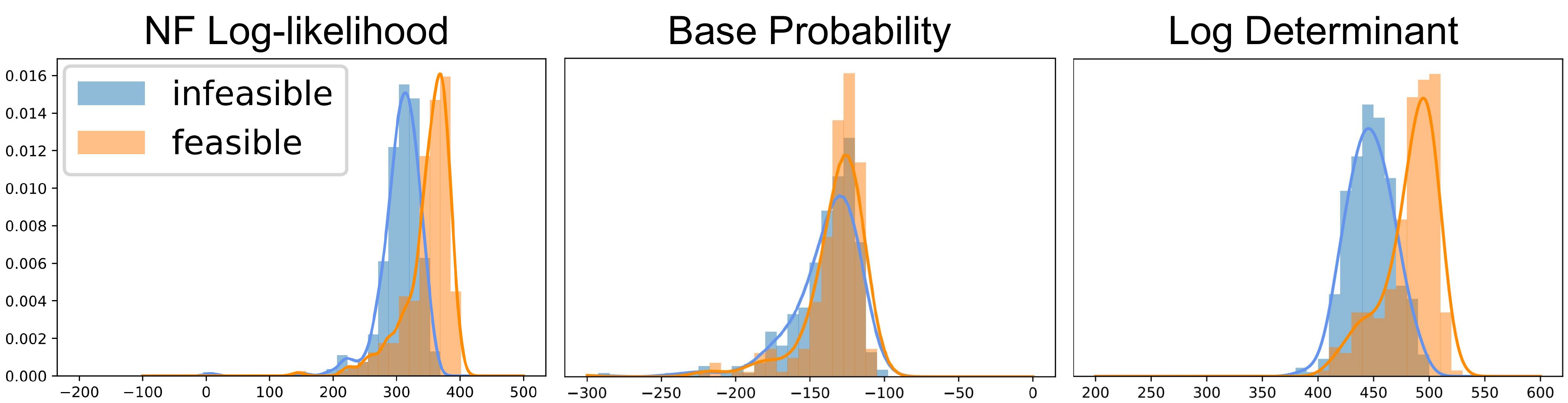}
    \caption{\textbf{\gls{nf} log-likelihoods for feasible and infeasible assemblies} with 5-parts (left), is a sum of the base probability (middle) and the transformation matrices log determinant (right). Best viewed in color.}
    \label{fig:log_likelihood}
\end{figure}

\paragraph{Likelihoods Ablation}The \gls{nf} log-likelihood estimation in Eq.~\ref{eq:mle} is a sum of two terms: the density of the base distribution and the log-determinant of the Jacobian of the flow transformation. 
To understand the contribution of each of these to the model's estimation, we plot their values separately for the model with Gaussian base distribution in Fig.~\ref{fig:log_likelihood}. 
As expected, the determinants are the main contributing factor to the final scores, whereas the values produced by the base distribution act as a normalization term.

\begin{figure}
    \centering
        \includegraphics[width=1\linewidth]{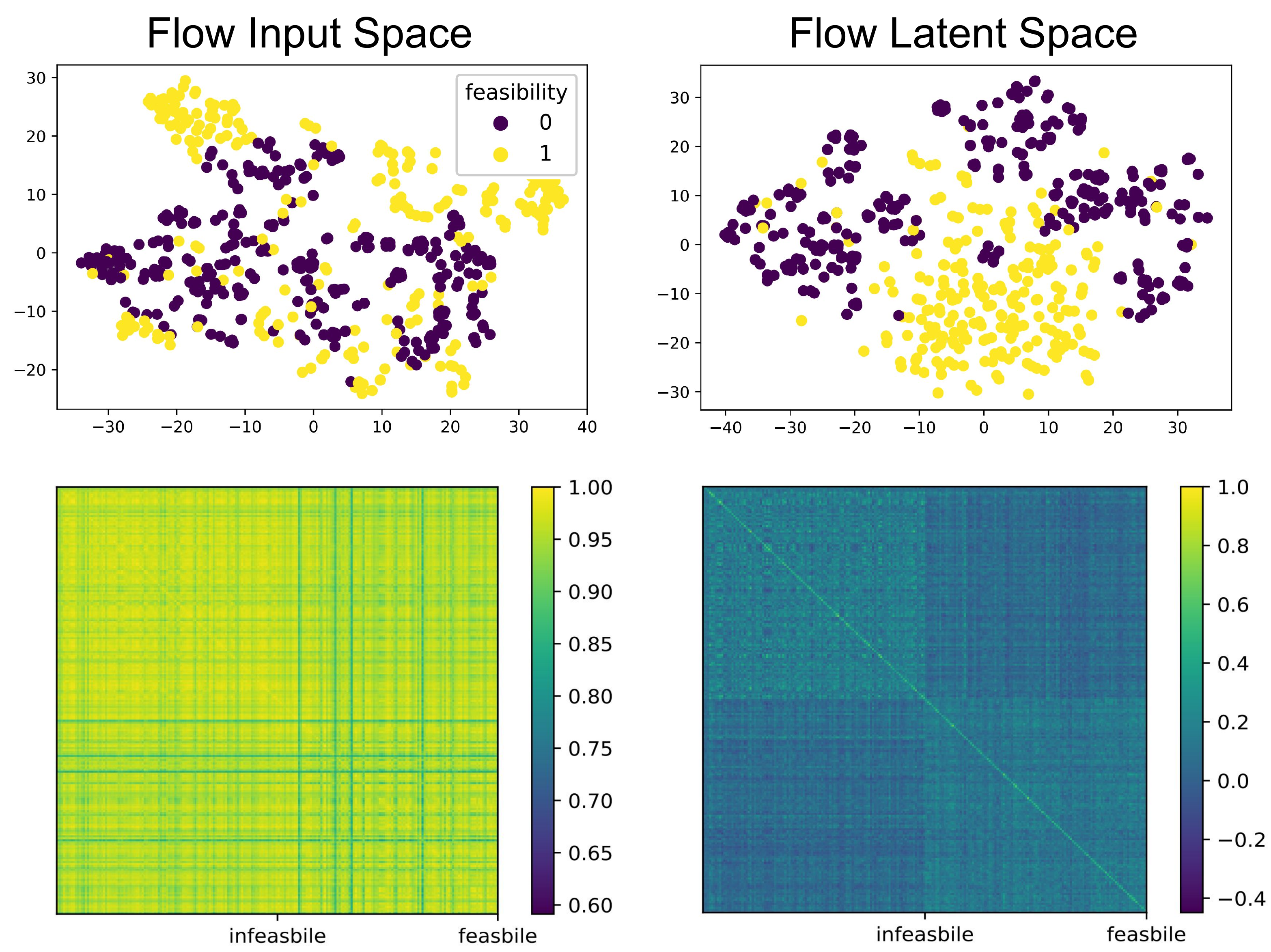}
    \caption{\textbf{Samples visualization in \gls{nf} input (left) and latent (right) spaces}. At the top, \gls{tsne} shows that samples mapped by \gls{nf} are "normalized", pulled together to a compact cluster. 
    At the bottom, \textit{Cosine Similarities} between feasible and infeasible assemblies are more distinct after the transformation, verifying the "normalization". Best viewed in color.}
    \label{fig:visualization}
\end{figure}

\paragraph{Samples Visualization} We visualize the coordinates of the embeddings in the input space (as created by the \gls{grace} feature extractor) and in the \gls{nf} latent space with \gls{tsne} and similarity matrices (Fig.~\ref{fig:visualization}). 
As shown in the \gls{tsne} visualization, the samples of feasible assemblies are pulled together and hence clustered more compactly when compared to those in the input space before the flow transformation.
This is verified again in the similarity matrices at the bottom, where the distances between feasible samples are smaller than those of infeasible ones after. 
These results show us that the flow transformation indeed \textit{"normalizes"} the inputs in terms of both likelihood computation and geometrical coordinates.
This observation also confirms the finding of better \gls{ood} detection performance in the flow latent space~\cite{jiang2022revisiting}, which is worth exploring for more effective feasibility learning algorithms, which we leave for future work. 
Besides, a further improvement could be archived by encouraging the feature extractor \gls{grace} to grasp semantics that are more closely related to the feasibility task, as suggested by~\cite{kirichenko2020normalizing}.

\section{Conclusion}

In this work, we seek to address feasibility prediction for data-driven methods in \gls{rasp} with \gls{nf} relying only on feasible examples.
With the formulation of density-based \gls{ood} detection, we develop an effective feasibility prediction algorithm based on feature embeddings from a pre-trained processing network. 
The empirical experiments on detecting infeasible assemblies in simulation present promising results, which outperform the baselines. 
We further dug into the internal working mechanism of \gls{nf} for this use case and found insightful observations, which can provide more understanding to inspire other researchers for further improvements in this direction.
For future research, we suggest introducing explainability into this setting with a gradient map in respect to the input, which can guide the user in altering the structure and enable its assembly, i.e., counter-factual explanation \cite{atad2022chexplaining}.

\section*{Acknowledgments}
We thank the anonymous reviewers for their thoughtful feedback. Jianxiang Feng is supported by the Munich School for Data Science (MUDS). Rudolph Triebel and Stephan Günnemann are members of MUDS.

\bibliographystyle{plainnat}
\bibliography{references}

\begin{thebibliography}{23}
\providecommand{\natexlab}[1]{#1}
\providecommand{\url}[1]{\texttt{#1}}
\expandafter\ifx\csname urlstyle\endcsname\relax
  \providecommand{\doi}[1]{doi: #1}\else
  \providecommand{\doi}{doi: \begingroup \urlstyle{rm}\Url}\fi

\bibitem[Atad et~al.(2022)Atad, Dmytrenko, Li, Zhang, Keicher, Kirschke,
  Wiestler, Khakzar, and Navab]{atad2022chexplaining}
Matan Atad, Vitalii Dmytrenko, Yitong Li, Xinyue Zhang, Matthias Keicher, Jan
  Kirschke, Bene Wiestler, Ashkan Khakzar, and Nassir Navab.
\newblock Chexplaining in style: Counterfactual explanations for chest x-rays
  using stylegan.
\newblock \emph{arXiv preprint arXiv:2207.07553}, 2022.

\bibitem[Atad et~al.(2023)Atad, Feng, Rodr{\'\i}guez, Durner, and
  Triebel]{atad2023efficient}
Matan Atad, Jianxiang Feng, Ismael Rodr{\'\i}guez, Maximilian Durner, and
  Rudolph Triebel.
\newblock Efficient and feasible robotic assembly sequence planning via graph
  representation learning.
\newblock \emph{arXiv preprint arXiv:2303.10135}, 2023.

\bibitem[Charpentier et~al.(2020)Charpentier, Z{\"u}gner, and
  G{\"u}nnemann]{charpentier2020posterior}
Bertrand Charpentier, Daniel Z{\"u}gner, and Stephan G{\"u}nnemann.
\newblock Posterior network: Uncertainty estimation without ood samples via
  density-based pseudo-counts.
\newblock \emph{Advances in Neural Information Processing Systems},
  33:\penalty0 1356--1367, 2020.

\bibitem[Dinh et~al.(2014)Dinh, Krueger, and Bengio]{dinh2014nice}
Laurent Dinh, David Krueger, and Yoshua Bengio.
\newblock Nice: Non-linear independent components estimation.
\newblock \emph{arXiv preprint arXiv:1410.8516}, 2014.

\bibitem[Dinh et~al.(2016)Dinh, Sohl-Dickstein, and Bengio]{dinh2016density}
Laurent Dinh, Jascha Sohl-Dickstein, and Samy Bengio.
\newblock Density estimation using real nvp.
\newblock \emph{arXiv preprint arXiv:1605.08803}, 2016.

\bibitem[Driess et~al.(2020)Driess, Oguz, Ha, and Toussaint]{driess2020deep}
Danny Driess, Ozgur Oguz, Jung-Su Ha, and Marc Toussaint.
\newblock Deep visual heuristics: Learning feasibility of mixed-integer
  programs for manipulation planning.
\newblock In \emph{2020 IEEE ICRA}, pages 9563--9569. IEEE, 2020.

\bibitem[Feng et~al.(2022)Feng, Durner, M{\'a}rton, B{\'a}lint-Bencz{\'e}di,
  and Triebel]{feng2022introspective}
Jianxiang Feng, Maximilian Durner, Zolt{\'a}n-Csaba M{\'a}rton, Ferenc
  B{\'a}lint-Bencz{\'e}di, and Rudolph Triebel.
\newblock Introspective robot perception using smoothed predictions from
  bayesian neural networks.
\newblock In \emph{Robotics Research: The 19th International Symposium ISRR},
  pages 660--675. Springer, 2022.

\bibitem[Jiang et~al.(2022)Jiang, Sun, and Yu]{jiang2022revisiting}
Dihong Jiang, Sun Sun, and Yaoliang Yu.
\newblock Revisiting flow generative models for out-of-distribution detection.
\newblock In \emph{ICLR}, 2022.

\bibitem[Kirichenko et~al.(2020)Kirichenko, Izmailov, and
  Wilson]{kirichenko2020normalizing}
Polina Kirichenko, Pavel Izmailov, and Andrew~G Wilson.
\newblock Why normalizing flows fail to detect out-of-distribution data.
\newblock \emph{Advances in neural information processing systems},
  33:\penalty0 20578--20589, 2020.

\bibitem[Lee et~al.(2020)Lee, Humt, Feng, and Triebel]{pmlr-v119-lee20b}
Jongseok Lee, Matthias Humt, Jianxiang Feng, and Rudolph Triebel.
\newblock Estimating model uncertainty of neural networks in sparse information
  form.
\newblock In Hal~Daumé III and Aarti Singh, editors, \emph{Proceedings of the
  37th ICML}, volume 119 of \emph{Proceedings of Machine Learning Research},
  pages 5702--5713. PMLR, 13--18 Jul 2020.

\bibitem[Ma et~al.(2022)Ma, Gong, Xu, Chen, Zhao, Huang, and
  Zhou]{ma2022planning}
Lin Ma, Jiangtao Gong, Hao Xu, Hao Chen, Hao Zhao, Wenbing Huang, and Guyue
  Zhou.
\newblock Planning assembly sequence with graph transformer.
\newblock \emph{arXiv preprint arXiv:2210.05236}, 2022.

\bibitem[Rodr{\i}guez et~al.(2019)Rodr{\i}guez, Nottensteiner, Leidner,
  Ka{\ss}ecker, Stulp, and Albu-Sch{\"a}ffer]{rodriguez2019iteratively}
Ismael Rodr{\i}guez, Korbinian Nottensteiner, Daniel Leidner, Michael
  Ka{\ss}ecker, Freek Stulp, and Alin Albu-Sch{\"a}ffer.
\newblock Iteratively refined feasibility checks in robotic assembly sequence
  planning.
\newblock \emph{IEEE RAL}, 4\penalty0 (2):\penalty0 1416--1423, 2019.

\bibitem[Rodr{\'\i}guez et~al.(2020)Rodr{\'\i}guez, Nottensteiner, Leidner,
  Durner, Stulp, and Albu-Sch{\"a}ffer]{rodriguez2020pattern}
Ismael Rodr{\'\i}guez, Korbinian Nottensteiner, Daniel Leidner, Maximilian
  Durner, Freek Stulp, and Alin Albu-Sch{\"a}ffer.
\newblock Pattern recognition for knowledge transfer in robotic assembly
  sequence planning.
\newblock \emph{IEEE RAL}, 5\penalty0 (2):\penalty0 3666--3673, 2020.

\bibitem[Rudolph et~al.(2021)Rudolph, Wandt, and Rosenhahn]{rudolph2021same}
Marco Rudolph, Bastian Wandt, and Bodo Rosenhahn.
\newblock Same same but differnet: Semi-supervised defect detection with
  normalizing flows.
\newblock In \emph{Proceedings of the IEEE/CVF WACV}, pages 1907--1916, 2021.

\bibitem[Sch{\"o}lkopf et~al.(1999)Sch{\"o}lkopf, Williamson, Smola,
  Shawe-Taylor, and Platt]{scholkopf1999support}
Bernhard Sch{\"o}lkopf, Robert~C Williamson, Alex Smola, John Shawe-Taylor, and
  John Platt.
\newblock Support vector method for novelty detection.
\newblock \emph{Advances in neural information processing systems}, 12, 1999.

\bibitem[Sinha et~al.(2022)Sinha, Sharma, Banerjee, Lew, Luo, Richards, Sun,
  Schmerling, and Pavone]{sinha2022system}
Rohan Sinha, Apoorva Sharma, Somrita Banerjee, Thomas Lew, Rachel Luo,
  Spencer~M Richards, Yixiao Sun, Edward Schmerling, and Marco Pavone.
\newblock A system-level view on out-of-distribution data in robotics.
\newblock \emph{arXiv preprint arXiv:2212.14020}, 2022.

\bibitem[Stimper et~al.(2022)Stimper, Sch{\"o}lkopf, and
  Hern{\'a}ndez-Lobato]{stimper2022resampling}
Vincent Stimper, Bernhard Sch{\"o}lkopf, and Jos{\'e}~Miguel
  Hern{\'a}ndez-Lobato.
\newblock Resampling base distributions of normalizing flows.
\newblock In \emph{AISTATS}, pages 4915--4936. PMLR, 2022.

\bibitem[Stimper et~al.(2023)Stimper, Liu, Campbell, Berenz, Ryll,
  Sch{\"o}lkopf, and Hern{\'a}ndez-Lobato]{normflows}
Vincent Stimper, David Liu, Andrew Campbell, Vincent Berenz, Lukas Ryll,
  Bernhard Sch{\"o}lkopf, and Jos{\'e}~Miguel Hern{\'a}ndez-Lobato.
\newblock normflows: {A} {P}y{T}orch {P}ackage for {N}ormalizing {F}lows.
\newblock \emph{arXiv preprint arXiv:2302.12014}, 2023.

\bibitem[Wells et~al.(2019)Wells, Dantam, Shrivastava, and
  Kavraki]{wells2019learning}
Andrew~M Wells, Neil~T Dantam, Anshumali Shrivastava, and Lydia~E Kavraki.
\newblock Learning feasibility for task and motion planning in tabletop
  environments.
\newblock \emph{IEEE RAL}, 4\penalty0 (2):\penalty0 1255--1262, 2019.

\bibitem[Xu et~al.(2022)Xu, Ren, Chalvatzaki, and Peters]{xu2022accelerating}
Lei Xu, Tianyu Ren, Georgia Chalvatzaki, and Jan Peters.
\newblock Accelerating integrated task and motion planning with neural
  feasibility checking.
\newblock \emph{arXiv preprint arXiv:2203.10568}, 2022.

\bibitem[Yang et~al.(2022)Yang, Garrett, and Fox]{yang2022sequence}
Zhutian Yang, Caelan~Reed Garrett, and Dieter Fox.
\newblock Sequence-based plan feasibility prediction for efficient task and
  motion planning.
\newblock \emph{arXiv preprint arXiv:2211.01576}, 2022.

\bibitem[Zhang et~al.(2020)Zhang, Li, Guo, and Guo]{zhang2020hybrid}
Hongjie Zhang, Ang Li, Jie Guo, and Yanwen Guo.
\newblock Hybrid models for open set recognition.
\newblock In \emph{Computer Vision--ECCV 2020, Proceedings, Part III 16}, pages
  102--117. Springer, 2020.

\bibitem[Zhao et~al.(2020)Zhao, Guo, Fang, and Ou]{Zhao2019}
M.~Zhao, X.~Guo, X.and~Zhang, Y.~Fang, and Y.~Ou.
\newblock Aspw-drl: assembly sequence planning for workpieces via a deep
  reinforcement learning approach.
\newblock \emph{Assembly Automation}, 40:\penalty0 65--75, 2020.
\newblock ISSN 0144-5154.

\end{thebibliography}

\end{document}